\documentclass[a4paper,10pt,fleqn,usenames,dvipsnames]{article}

    \usepackage{lmodern}
\usepackage[T1]{fontenc}
\usepackage{amsmath}
\usepackage{amssymb}

\usepackage{url}
\usepackage{longtable}
\usepackage{multirow}
\usepackage{caption}
\usepackage[labelformat=simple]{subcaption}

\usepackage[nodayofweek]{datetime}
\usepackage{enumerate}
\usepackage{xspace}
\usepackage{xcolor}
\usepackage{graphicx}
\usepackage[pdftex,hidelinks]{hyperref}
\usepackage{bookmark}

\usepackage{float}
\usepackage[capitalise]{cleveref}

\usepackage{chngcntr}
\usepackage{placeins}

\usepackage[backend=biber,giveninits=true,doi=false,sorting=none,style=numeric-comp]{biblatex}
\DeclareFieldFormat[article]{citetitle}{\textit{#1}\isdot}
\DeclareFieldFormat[article]{title}{\textit{#1}\isdot}
\DeclareFieldFormat[unpublished]{citetitle}{\textit{#1}\isdot}
\DeclareFieldFormat[unpublished]{title}{\textit{#1}\isdot}
\DeclareFieldFormat[inproceedings]{citetitle}{\textit{#1}\isdot}
\DeclareFieldFormat[inproceedings]{title}{\textit{#1}\isdot}

\DeclareNameAlias{author}{last-first}

\usepackage[nonatbib,preprint]{neurips}

\title{Flows for Flows: Training Normalizing Flows Between Arbitrary Distributions with Maximum Likelihood Estimation}
\author{%
  Samuel Klein$^\dagger$ \\
  University of Geneva\\
  \texttt{samuel.klein@unige.ch} \\
  \And
  John Andrew Raine$^\dagger$ \\
  University of Geneva\\
  \texttt{john.raine@unige.ch} \\
  \And
  Tobias Golling \\
  University of Geneva\\ 
  \texttt{tobias.golling@unige.ch} \\
}

\newcommand{\fff}{{Flows for Flows}\xspace}

\newcommand{\fffcode}{\texttt{ffflows}\xspace}

\begin{document}
\maketitle

    \begin{abstract}
          Normalizing flows are constructed from a base distribution with a known density and a diffeomorphism with a tractable Jacobian.
          The base density of a normalizing flow can be parameterised by a different normalizing flow, thus allowing maps to be found between arbitrary distributions.
          We demonstrate and explore the utility of this approach and show it is particularly interesting in the case of conditional normalizing flows and for introducing optimal transport constraints on maps that are constructed using normalizing flows.
          \vfill
          $^\dagger$ Contributed equally to this paper and the \texttt{flows4flows} package.
    \end{abstract}

    \section{Introduction}
A normalizing flow~\cite{tabak_flows,flows_review} is a popular density estimator in the field of machine learning. 
These models are most often used as generative models, with some applications in estimating densities.
A normalizing flow is defined by an invertible map between an arbitrary distribution and a base distribution with a known density. The most common choice is a standard normal distribution.

In this paper we explore the use of normalizing flows for learning maps between two arbitrary distributions whose probability distribution functions need not be known a priori.

Learning maps between different distributions is a useful task for performing calibration, but also for doing interpolation between distributions or across conditional densities as is required in many domains including high energy physics~\cite{hallin2021classifying,Andreassen:2020nkr,curtains,naf_datatransform}. Furthermore by moving between arbitrary distributions a flow can be constructed to learn simpler transformations between states which could have various applications in diffusion modelling or interpolation.
An additional application, though not studied in detail in this work, is further investigation of the performance of models trained directly on discrete distributions, and how they compare to other approaches generalising normalizing flows to discrete distributions.


The possibility of using normalizing flows as maps between different distributions is known in the wider community but to date has not been described or explored in detail.

In this work we demonstrate the capabilities of normalizing flows as maps between distributions and explore approaches for regularizing these maps to provide semantic meaning to the learned transformation.
We focus on 2-dimensional toy distributions to visually demonstrate the performance in an extreme but domain unspecific setting. 
However, applications of normalizing flows such as those presented in Refs.~\cite{curtains,naf_datatransform} could greatly benefit from training with exact maximum likelihoods instead of distance losses.

We will refer to the training of flows between distributions as \fff.


First, we study the simplest construction where we apply \fff to learn a transport map between two completely independent distributions. Here we demonstrate the flexibility of normalizing flows and compare the performance to using two separate normalizing flows.

Second, we study the application of \fff on conditional distributions. 
The normalizing flow is trained to map between distributions sampled from the same probability distribution $p(x ; c)$ but for different values of $c$. 
In this application the \fff model learns the conditional transport map along the joint distribution.

The combination of both of these first two studies is a trivial extension, whereby one learns to map between two arbitrary distributions with a common conditional value.
This is not studied in this work, however could be very useful as an alternative to reweighting and resampling approaches.

Third, we study the consequences and benefits of using \fff in comparison to two normalizing flows with a shared base distribution, with a focus on introducing a distance penalty in the loss function.
This is a simplified approach to encourage the normalizing flow to learn an optimal transport map. In principle a model such as \texttt{CP-Flows}~\cite{cp_flows} can be used in conjunction with \fff to define the transformer between the two distributions.

In particular we present a new package \fffcode\footnote{\url{https://github.com/jraine/flows4flows/}}, which simplifies the construction of normalizing flows between two arbitrary (conditional) distributions using the \texttt{nflows}~\cite{nflows} package as a foundation.

    \section{Background}

Normalizing flows are defined by a parameteric diffeomorphism $f_\phi$ and a base density $p_\theta$ for which the density is known.
Using the change of variables formula the log likelihood of a data point $x$ under a normalizing flow is given by
\begin{equation}
    \log p_{\theta, \phi} (x) = \log p_\theta (f_\phi^{-1}(x)) - \log \left| \det (J_{f_\phi^{-1}(x)}) \right|,
\end{equation}
where $J$ is the Jacobian of $f_\phi$.
Training the model to maximise the likelihood of data samples results in a map $f_\phi^{-1}$ between the data distribution $p_D(x)$ and the base density $p_\theta$.

The base density of a normalizing flow can be any distribution for which we know the density, including another normalizing flow that has been trained on an arbitrary dataset.
This means normalizing flows can be trained to find maps between two different distributions. 
Using a normalizing flow as the base density is referred to in this paper as a flow for flow.

\begin{figure}
    \centering
    \includegraphics[]{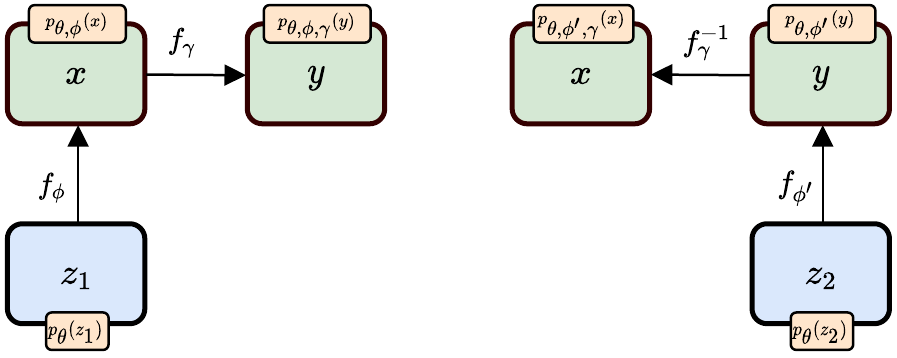}
    \caption{A schematic of the flows for flows architecture.}
    \label{fig:fff_schematic}
\end{figure}

Given samples from two data distributions $x \sim p_{D_1}(x)$ and $y \sim p_{D_2}(y)$ of the same dimensionality a map $f_\gamma: x \mapsto y$ between these distributions can be found by inducing a density on $X$ to use as the base density in the construction of a normalizing flow. 
A density can be induced on $X$ by defining a new space $Z$ with base distribution $p_\theta(z)$ and an invertible function $f_\phi: z \mapsto x$.
The respective densities can then be fit to the data by maximising the log likelihood of the data under the densities defined by the change of variables formula and given by
\begin{align*}
    \begin{split}
    \max_\gamma \mathop{\mathbb{E}}_{y \sim p_{D_2}} \left[ \log p_{\theta, \phi, \gamma}(y) \right]
    &= \max_\gamma \mathop{\mathbb{E}}_{y \sim p_{D_2}} \left[ \log p_{\theta, \phi}(f_\gamma^{-1}(y)) - \log \left| \det (J_{f_\gamma^{-1}(y)}) \right| \right], \\
    \max_\phi \mathop{\mathbb{E}}_{x \sim p_{D_1}} \left[ \log p_{\theta, \phi}(f_\gamma^{-1}(y)) \right]
    &= \max_\phi \mathop{\mathbb{E}}_{x \sim p_{D_1}} \left[ \log p_{\theta}(f_\phi^{-1}(x)) - \log \left| \det (J_{f_\phi^{-1}(x)}) \right| \right].
    \end{split}
\end{align*}
Choosing to induce a density on $X$ is arbitrary, and the same construction can be performed by inducing a density on $Y$. 
Defining densities on both $X$ and $Y$ allows both $f_\gamma$ and $f_\gamma^{-1}$ to be used during training.
A benefit of training in both directions is that the dependance of $f_\gamma$ on the base densities on $X$ and $Y$ is reduced.
A schematic of the flow for flow architecture with densities on both $X$ and $Y$ is shown in Fig.~\ref{fig:fff_schematic}.

The invertible network $f_\gamma$ that is used to map between the two distributions has no semantic meaning on its own as some invertible neural networks are known to be universal function approximators. 
This map can become interpretable if it is subject to additional constraints, such as minimizing the transport distance between the two distributions.
Such a function can be learned by defining $f_\gamma$ to be the gradient of a convex potential as shown in \texttt{CP-Flows}~\cite{cp_flows}.

\paragraph*{Conditional normalizing flows}
Normalizing flows can learn conditional distributions by making the parameters of the invertible neural network dependent on the condition. The log-likelihood for a normalizing flow conditioned on some variables $c$ is defined by
\begin{equation}
    \log p_{\theta, \phi} (x | c) = \log p_\theta (f_{\phi(c)}^{-1}(x) | c) - \log \left| \det (J_{f_{\phi(c)}^{-1}(x)}) \right|,
\end{equation}
where the base density is also conditionally dependent on $c$.
In the case of conditional distributions with continuous conditions the data distribution $p_D(x | c)$ changes smoothly as a function of the condition.
For these settings a flow that maps between two different conditioning values can be much simpler than learning a map to an arbitrary base distribution.
Also in some settings, such as the approach in High Energy Physics presented in Ref.~\cite{curtains}, it is desireable to do interpolation along a conditional distribution.
In these settings it is desirable to have a strongly regularized function to do the interpolation, and so learning to map along the conditional distribution becomes desirable.

\begin{figure}
    \centering
    \includegraphics[width=0.7\textwidth]{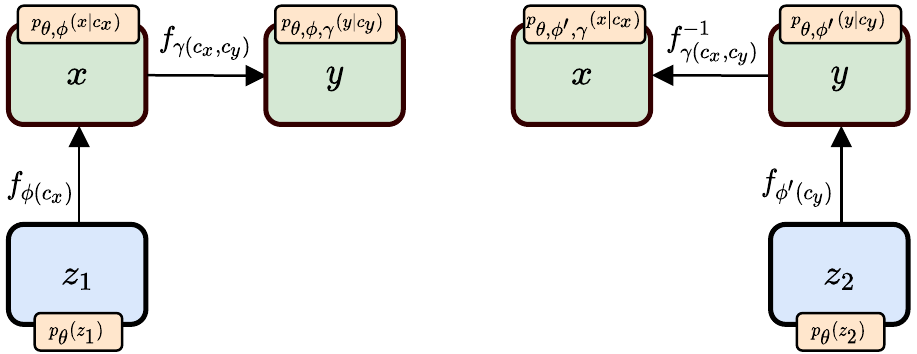}
    \caption{Schematic of a conditional flow for flow architecture.}
    \label{fig:schematic_flow4flow_conditional} 
\end{figure}

A schematic of a conditional flow for flow model is shown in Fig.~\ref*{fig:schematic_flow4flow_conditional} where the conditioning function $f_{\gamma(c_x, c_y)}$ can also take more restrictive forms, such as $f_{\gamma(c_x - c_y)}$ to ensure that the learned map is simple.
Further the two conditional base distributions can be identical such that $\phi = \phi'$.
Alternatively the base distributions can be different and instead a shared condition can be use $c=c_x=c_y$.



    \section{Flows for flows}

In this section we will demonstrate the flows for flows concept on complex low dimensional distributions.

All invertible neural networks are constructed from rational quadratic splines with four autoregressive layers~\cite{durkan2019neural}. 
Each spline transformation had eight bins and the parameters of the spline are defined using masked autoregressive networks with two blocks and $128$ nodes as defined in the \texttt{nflows} package~\cite{nflows}.
An initial learning rate of $10^{-4}$ was annealed to zero following a cosine schedule~\cite{cosine_annealing}. 
All trainings used a batch size of $128$ and the norm of the gradients were clipped to five.
In all cases $10^6$ points are sampled to form the training data.

\subsection{Unconditional distributions}
To demonstrate the efficacy of the flow for flow method we consider learning maps between arbitrary two dimensional datasets.
In this setting there is no clear benefit in using a flow for flow method, as evidenced by the similarity between using $f_\gamma$ and $f_{\phi'} \circ f_\phi^{-1}$.
This application does, however, demonstrate the ability of the \fff approach to learn maps between different distributions.
In this section all base densities are trained for ten epochs and the Flows For Flows model was trained for 20 epochs.

\begin{figure}
    \centering
    \includegraphics[width=0.75\textwidth]{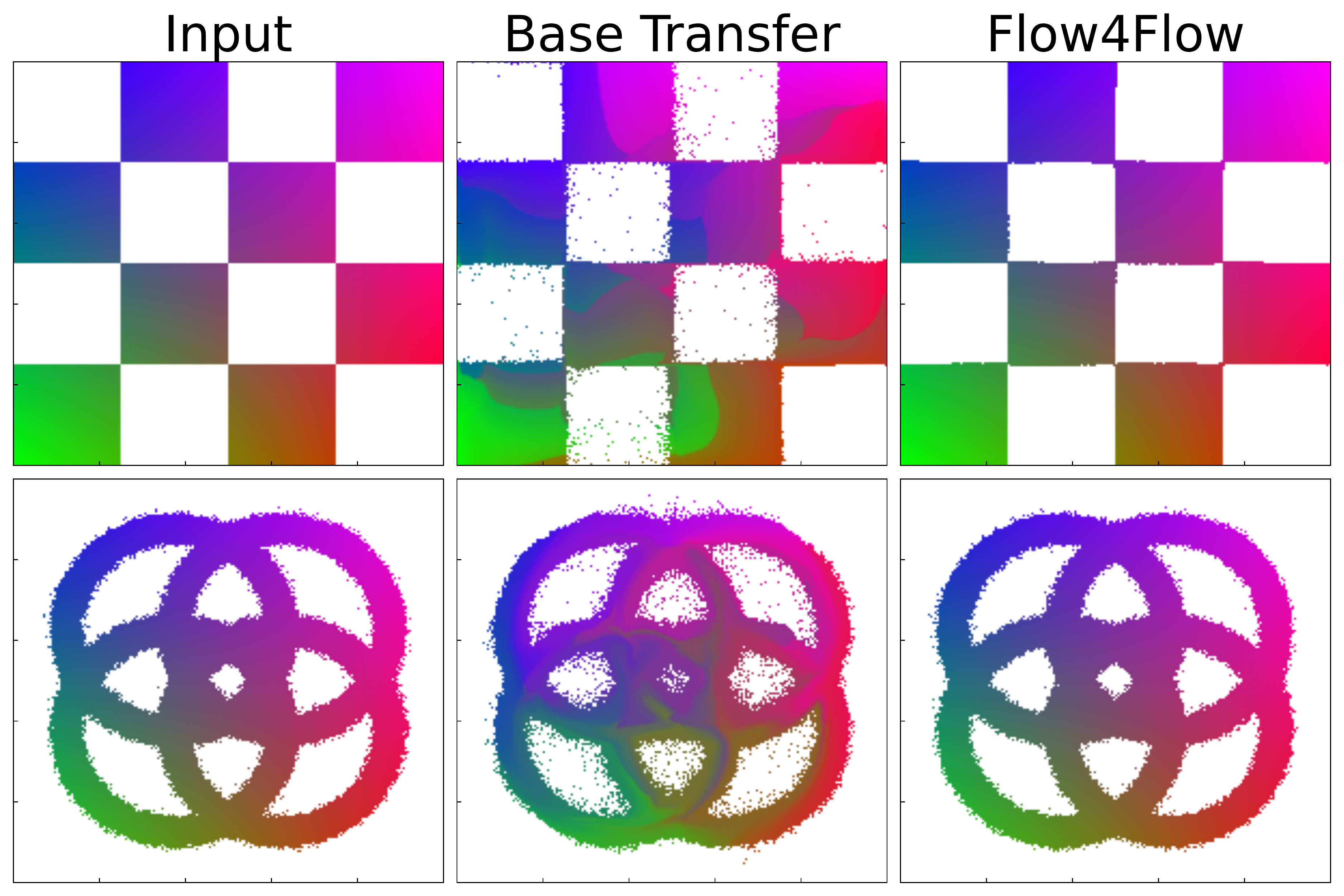}
    \caption{A flow for flow model trained to learn a map between two $2$D checkerboard (top) and four circle (bottom) distributions. Both models in this figure were trained using maximum likelihood with no regularisation. 
    A Base Transfer refers to using $f_{\phi'} \circ f_\phi^{-1}$.
    Each input point is assigned a color that is then matched to the output point, such that the distance between input and output points can be visualised.
    }
    \label{fig:identity_map} 
\end{figure}

\paragraph{Learning the identity}
In the simplest case the two data distributions between which we want to map are identical, and as demonstrated in Fig.~\ref*{fig:identity_map} this map can be learned with a flow for flow model.
Using the map $f_{\phi'} \circ f_\phi^{-1}$ is also possible in this setting, but as the base densities are trained separately the resulting map is very far from the identity as shown by the color coding of the input points in Fig.~\ref*{fig:identity_map}.
This could be remedied with joint training of the base densities, but in real world settings the two distributions are likely to be similar~\cite{curtains} and so mapping to an intermediate gaussian representation represents additional and unnecessary computation.
In contrast the flow for flow model learns a close approximation to the identity without any need for regularisation, though this appears to be closely linked to the initialisation of the invertible neural network.

\begin{figure}
    \centering
    \includegraphics[width=\textwidth]{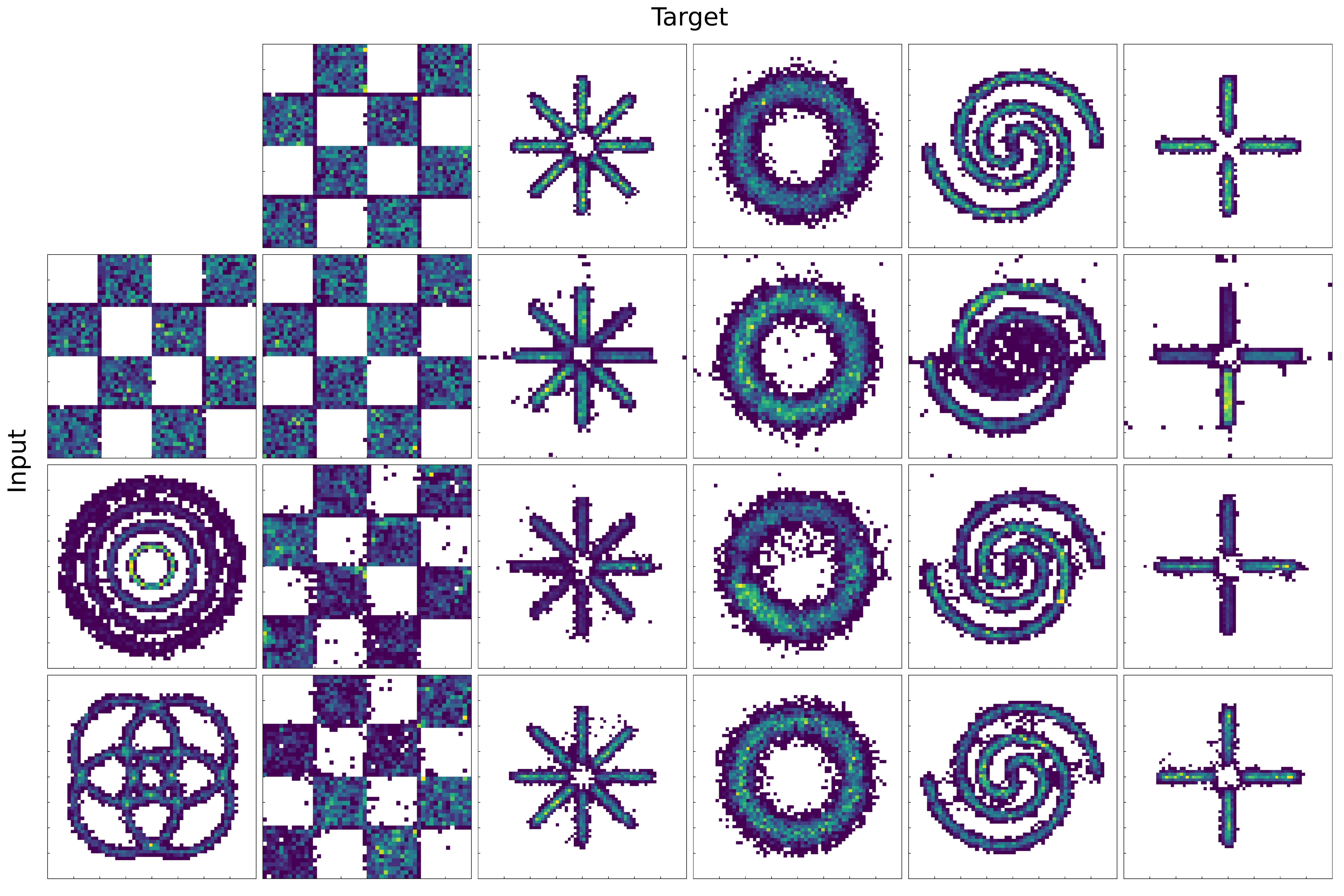}
    \caption{Flow for flow models trained to learn a map between various $2$D distributions. The models in this figure were trained using maximum likelihood with no regularisation. 
    }
    \label{fig:dist2dist_grid}
\end{figure}

\paragraph{Different distributions}
A more thorough test of using flows for flows is to learn the transformation between any pair of arbitrary distributions with the same dimensionality. 
In Fig.~\ref{fig:dist2dist_grid} flows for flows are used to transport between three input and four target distributions of varying complexity.
In all cases using the two base distributions directly outperforms the flows for flows, though it should be noted that it is not envisioned that flows for flows would find a practical application to such transformations and the base densities together have double the number of parameters as the Flows For Flows model.
The key point here is that the Flows For Flows approach can be used to find a map between arbitrary distributions, and the performance can be improved with more parameters (see Fig.~\ref{fig:dist2dist_grid_bigger}).

\subsection{Distance penalty}

As it is possible to regularise or constrain the normalizing flow used to transport data in \fff, we introduce an L1-norm distance penalty on the input and output data.
This distance regularisation is trivial to introduce for \fff but non-trivial in the case where two normalizing flows are used to move data from one distribution to another.

\begin{figure}
    \centering
    \includegraphics[width=\textwidth]{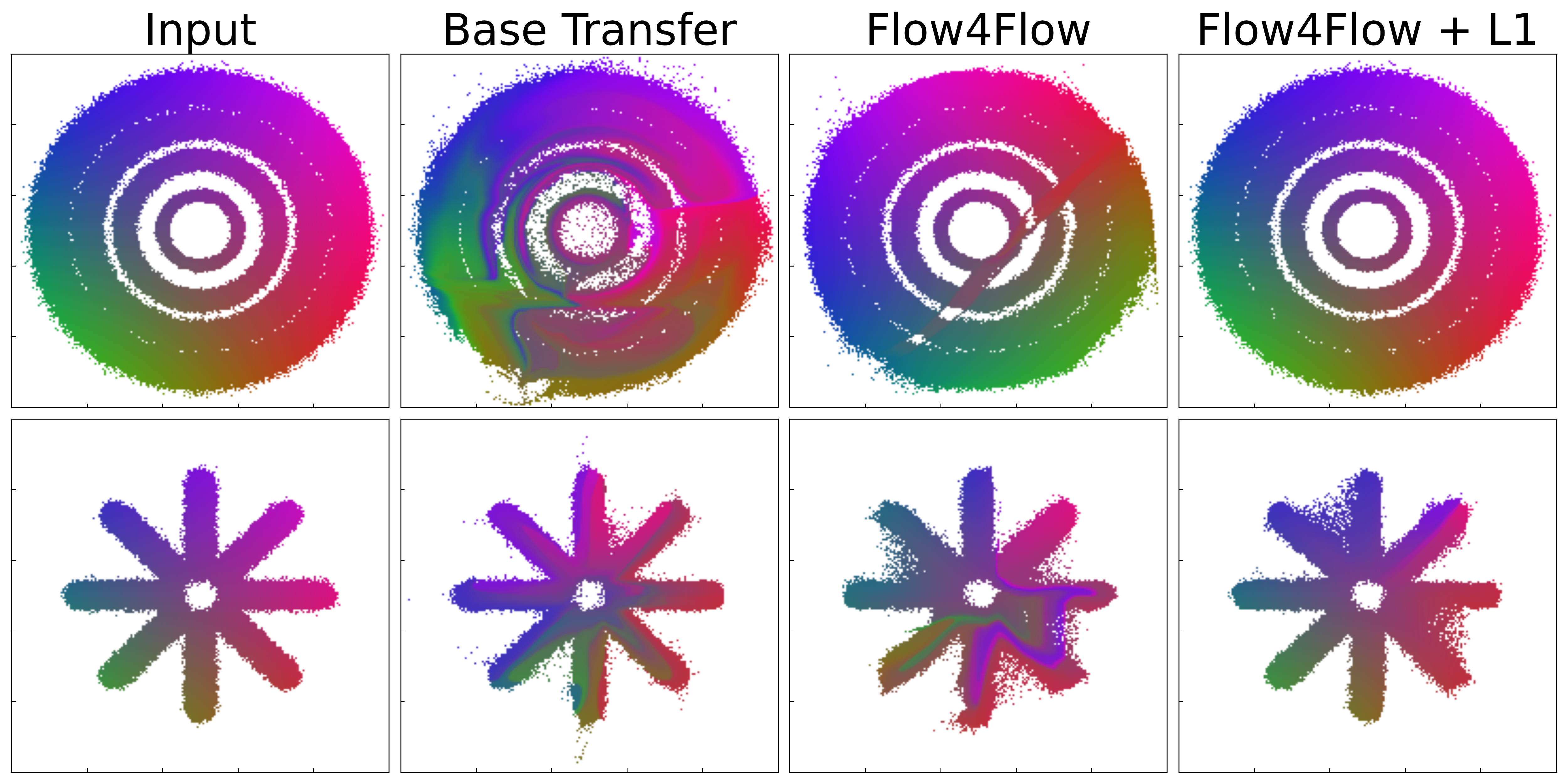}
    \caption{A flow for flow model trained to learn a map between two $2$D concentric ring (top) and two eight star (bottom)  distributions. The Flow For Flow models are trained using maximum likelihood with no regularisation (top) and L1 regularisation (bottom). 
    A Base Transfer refers to using $f_{\phi'} \circ f_\phi^{-1}$.
    Each input point is assigned a color that is then matched to the output point, such that the distance between input and output points can be seen directly.
    } 
    \label{fig:identity_map_pen} 
\end{figure}

To demonstrate its ability to better regularise the overall shift of data we introduce this to \fff trained to learn the identity and compare this to using two normalizing flows and transporting data via encoding to and decoding from the base distribution.
The values are shown in Tab.~\ref{tab:avgdist_identity}.
We can see that out of the box \fff typically learns not to greatly displace the data and learn the identity, however with the introduction of the distance penalty it is almost perfect in learning the identity function. The biggest impact is seen for the concentric circles distribution, as shown in Fig.~\ref{fig:identity_map_pen}.

\begin{table}[h]
    \centering
    \caption{Average translation distance of data transformed with a \fff model, a \fff model with an L1 distance penalty with a weight of 1.0, and by an encoding-decoding with the two normalizing flows used to train the \fff.
    The same distribution is used for both the input and output, but two separate normalizing flows are trained for the base distribution. All values are absolute mean translations.
    The datasets marked with $^{\dagger}$ have been run with a different random initialisation.
    Models trained with only the L1 norm distance penalty are shown in the last common to quantify the lower bound of expected performance.
    } 
    \begin{tabular}{ r | c c c | c}
        2D dataset & Flow4Flow & Flow4Flow + L1 & Base Density transport & L1 norm only\\
        \hline
        Ring & 0.216 & 0.002 & 0.566 & 0.001\\
        Concentric Rings & 2.184 & 0.003 & 1.108 & 0.001\\
        Four Circles & 0.006 & 0.003 & 1.363 & 0.001\\
        Checkerboard & 0.015 & 0.003 & 1.347 & 0.001\\
        Spirals & 0.185 & 0.003 & 1.410 & 0.002\\
        Star & 0.019 & 0.002 & 0.526 & 0.001\\
        Eight Star & 1.576 & 0.430 & 1.474 & 0.001\\
        \hline
        Ring$^{\dagger}$ & 0.025 & 0.002 & 0.679 & -- \\
        Concentric Rings$^{\dagger}$ & 2.823 & 0.008 & 2.174 & -- \\
        Eight Star$^{\dagger}$ & 0.516 & 0.002 & 1.019 & -- \\
    \end{tabular}
    \label{tab:avgdist_identity}
\end{table}

In general this is not of interest on its own, however it demonstrates the natural tendency of \fff to converge on a simpler transport map. Introducing an additional regularising term in the form of a distance penalty provides improved flexibility and better performance on functions that target the optimal transport map and settings where the input and target distributions are similar. 

\subsection{Conditional distributions} 

The distributions that have been studied so far can be extended by introducing rotations and radial scaling. 
A flow for flow model can then learn to move points sampled at one value of the condition such that they follow the distribution defined by another value of the condition. 
The flow for flow model will be conditioned on the difference between the two conditional values in all cases considered here.
All base densities in this section are trained for $32$ epochs and the Flows For Flows model is trained for $12$ epochs.

\begin{figure}[htbp]
    \centering 
    \includegraphics[width=\textwidth]{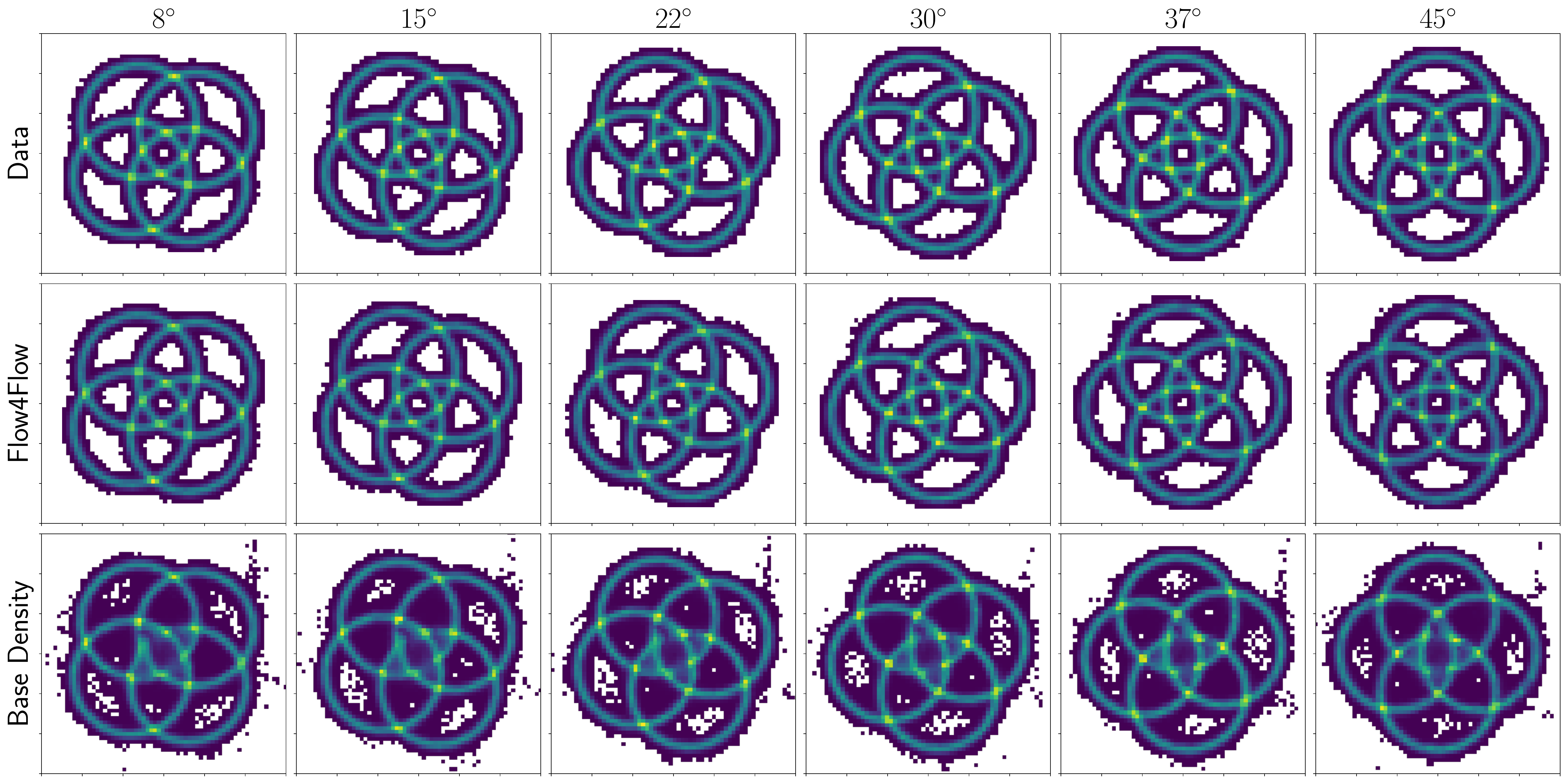}
    \caption{A flow for flow model trained on a distribution composed from four overlapping circles conditioned on the angle. 
    All models in this figure were trained using maximum likelihood with no regularisation.
    The flow for flow (Flow4Flow) model is given a four circles distribution with no rotation as input.
    }
    \label{fig:rotating_circles_density_compare} 
\end{figure}

The power of this approach is illustrated in Fig.~\ref{fig:rotating_circles_density_compare} where a flow for flow model is trained on a distribution composed of four overlapping circles which is conditionally rotated by up to $45$ degrees.
The benefit of a flow for flow model in this setting is that the map between the distributions defined at two different conditional values is continuous and so can be approximated by the invertible neural network that defines the flow for flow.
The map between distributions defined at different values of the condition is also simpler than the map between a standard normal and the conditional distribution.

In contrast the conditional base density has to learn to approximate a discontinuous map from a standard normal distribution to the data directly.
The invertible neural network that defines the base density is a continuous function and so cannot parameterise a discontinuous map.
This means the tails of the four circles distribution are poorly modelled by the base density and results in out of distributions samples as shown in Fig.~\ref{fig:rotating_circles_density_compare}.

\begin{figure}[htbp]
    \centering
    \includegraphics[width=\textwidth]{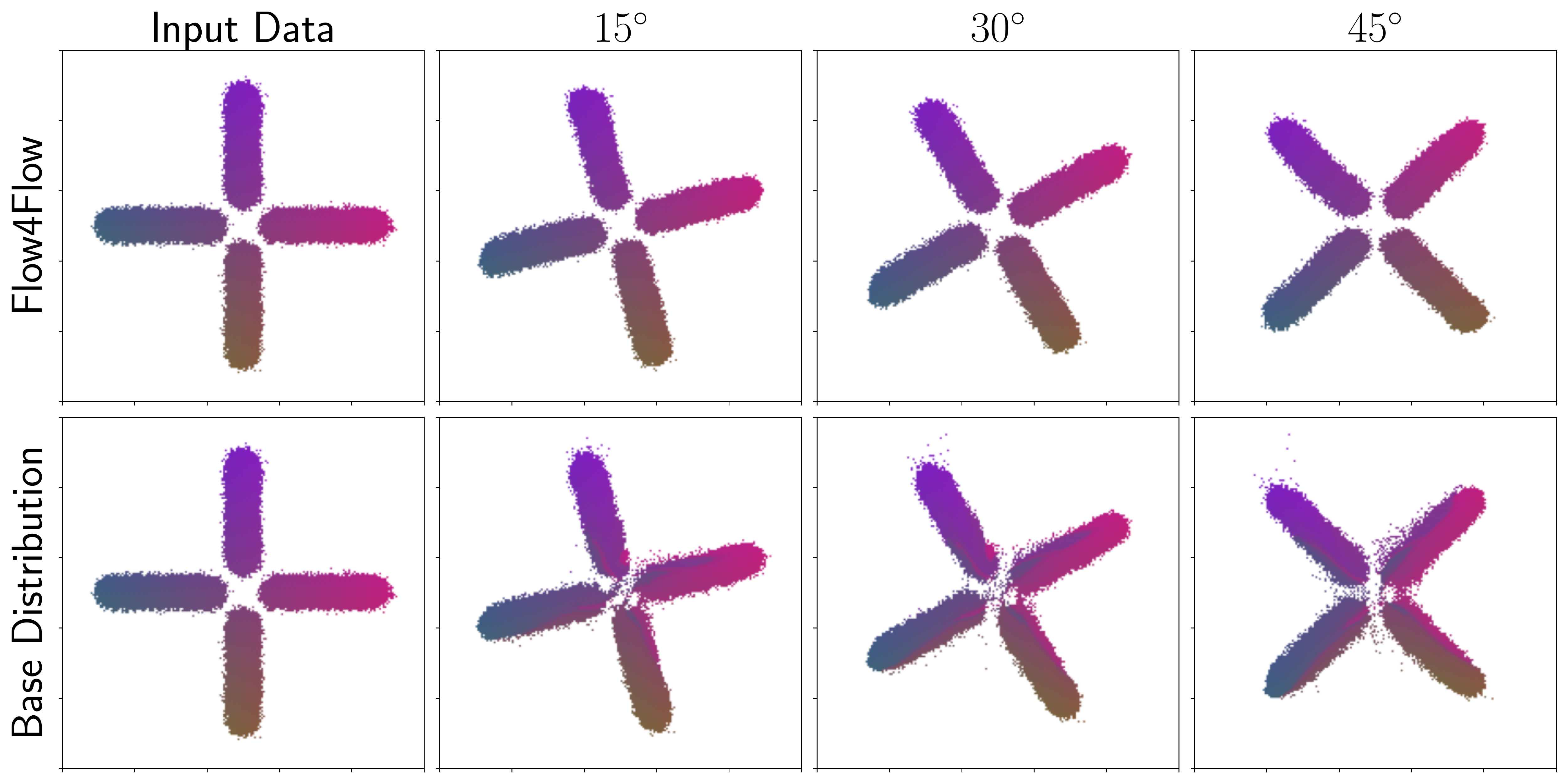}
    \caption{
        Color coded input data points are mapped to the same color output point for $15^\circ, 30^\circ$ and $45^\circ$ degree rotations using a Flow For Flow (Flow4Flow) and $f_{\phi'} \circ f_\phi^{-1}$ (Base Distribution).
    } 
    \label{fig:rotating_star_colored}
\end{figure}

Another way of mapping between two values of the condition is to go via the base density map $f_{\phi'} \circ f_\phi^{-1}$.
In the case of discontinuous distributions this again results in out of distribution samples as shown in Fig.~\ref{fig:rotating_star_colored}.
Further, by color coding the input data it can be seen that the $f_{\phi'} \circ f_\phi^{-1}$ does not have the same continuity as is inherent in the flows for flows model. 
A useful feature of the flows for flows approach is that the learned map has some semantic cohesion without any explicit regularisation.
This is however not guaranteed, which can be addressed by introducing an explicit optimal transport cost, though we find it has negligible impact in these settings.

The performance of the Flows For Flows approach is replicated in other conditional distributions as shown in App.~B.

    \FloatBarrier
    \section{Conclusions and future work}

    As demonstrated in this work it is possible to train a normalizing flow between any two arbitrary distributions of the same dimensionality. By using additional normalizing flows to parametrise the probability term, it is possible to train these normalizing flows with exact maximum likelihoods.

These models, which we call \fff, are capable of learning arbitrary maps, and could find many applications in domains relying on the modelling and transport of distributions. 
In particular they show great prospects in their use for learning transport maps on conditional probability distributions between different conditional values. Here they outperform sampling new data points from a conditional normalizing flow which have a higher fraction of out of distribution points.

By learning a mapping between two distributions it is possible to apply additional regularisation terms to the training, for example to minimise the distance of each individual shift in the data. 
This could be very interesting for applications trying to learn the optimal transport map between distributions. 
Furthermore, models such as those presented in \texttt{CP-Flows}.~\cite{cp_flows} could be used to define the network in \fff which should result in learning the optimal transport map between any two distributions. 

    \section*{Acknowledgements}
    The authors would like to acknowledge funding through the SNSF Sinergia grant called Robust Deep Density Models for High-Energy Particle Physics and Solar Flare Analysis (RODEM) with funding number CRSII$5\_193716$.
    
    \phantomsection
    \addcontentsline{toc}{chapter}{References}
    \printbibliography[title=References]
    
    \clearpage
    \appendix
    \counterwithin{figure}{section}
    \section{Arbitrary maps}

\subsection*{Increasing paramters of Flow for Flow}

\begin{figure}[h]
    \centering
    \includegraphics[width=\textwidth]{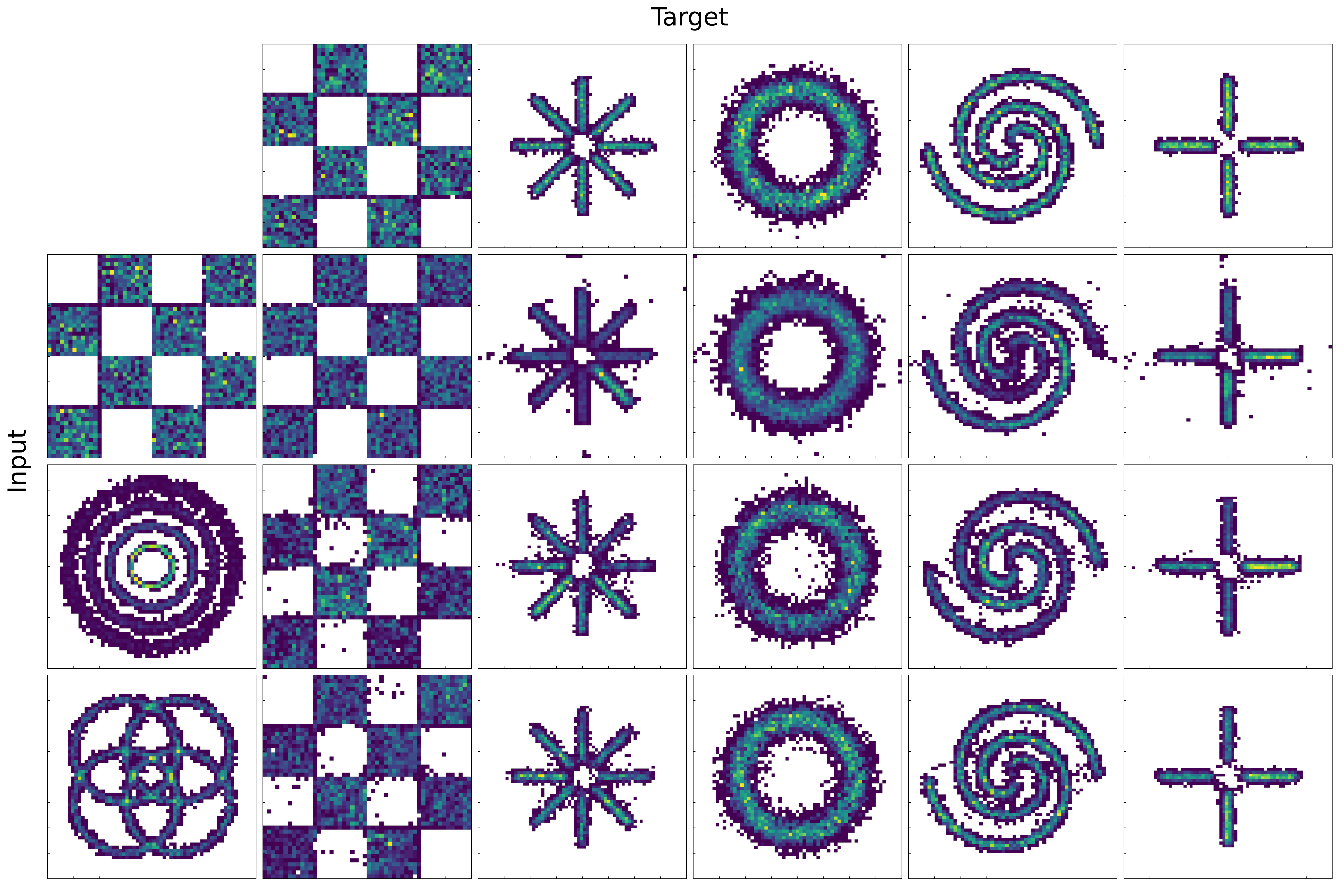}
    \caption{Flow for flow models trained to learn an identity map between various $2$D distributions with different base distributions for each side of the \fff, with additional L1 regularisation and without. 
    A Base Transfer refers to using $f_{\phi'} \circ f_\phi^{-1}$.
    Each input point is assigned a color that is then matched to the output point, such that the distance between input and output points can be seen directly.
    Each input point is assigned a color that is then matched to the output point, such that the distance between input and output points can be seen directly.
    }
    \label{fig:dist2dist_grid_bigger}
\end{figure}

By increasing the complexity of the Flow for Flow model, using an additional block for each rational quadratic spline and one additional layer, better overall performance can be seen in Fig.~\ref{fig:dist2dist_grid_bigger}. 

\subsection*{Additional identity maps}

\begin{figure}[h]
    \centering
    \includegraphics[width=\textwidth]{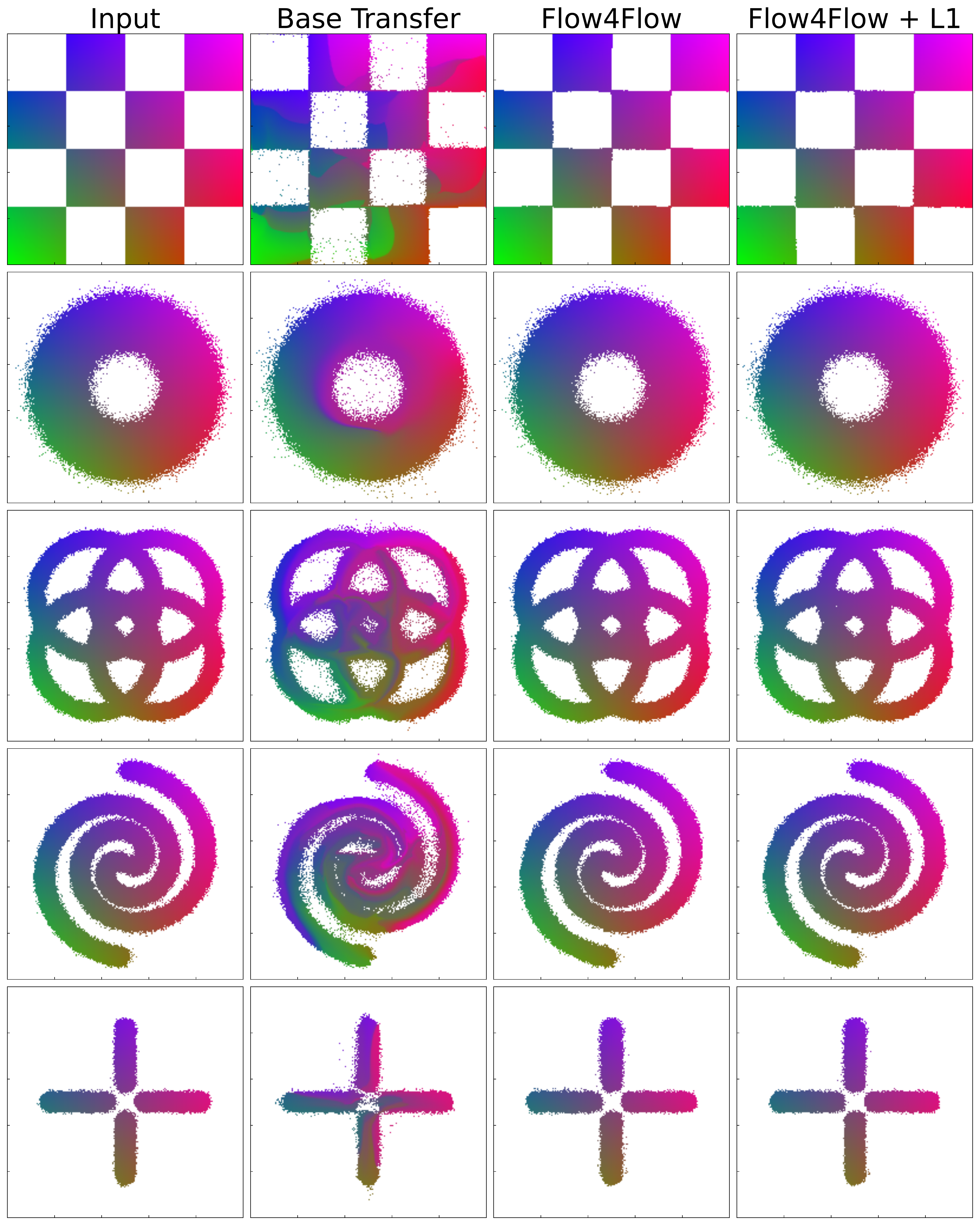}
    \caption{Flow for flow models trained to learn a map between various $2$D distributions. All models in this figure were trained using maximum likelihood with no regularisation.}
    \label{fig:morecolident}
\end{figure}

Colour maps for the remaining distributions considered are shown in Fig.~\ref{fig:morecolident}.

    \section{Conditional distributions}
\label{app:conditional_dists}

\subsection*{Rotating four circles}
\begin{figure}[h]
    \centering
    \includegraphics[width=\textwidth]{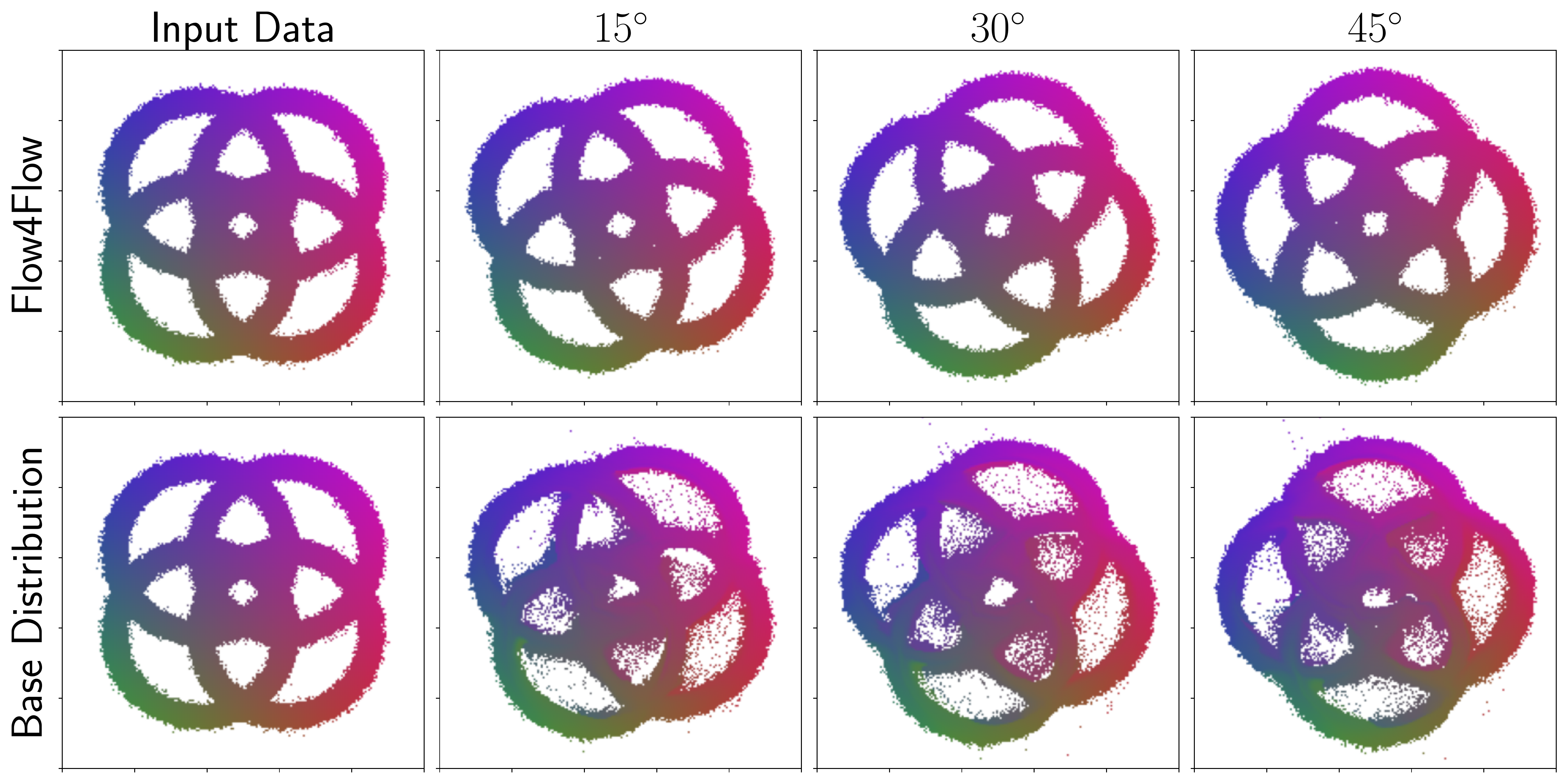}
    \caption{
        A color encoding of the rotating four circles distribution with a comparison between the performance of the flow for flow model and the base distribution map constructed using $f_{\phi'} \circ f_\phi^{-1}$.
    }
    \label{fig:colored_four_circles_rotate}
\end{figure}

In this section we provide the colored encoding map Fig.~\ref{fig:colored_four_circles_rotate} that corresponds to the densities that are shown in Fig.~\ref*{fig:rotating_circles_density_compare}.

\subsection*{Additional conditional distributions}

\begin{figure}[h]
    \centering 
    \includegraphics[width=\textwidth]{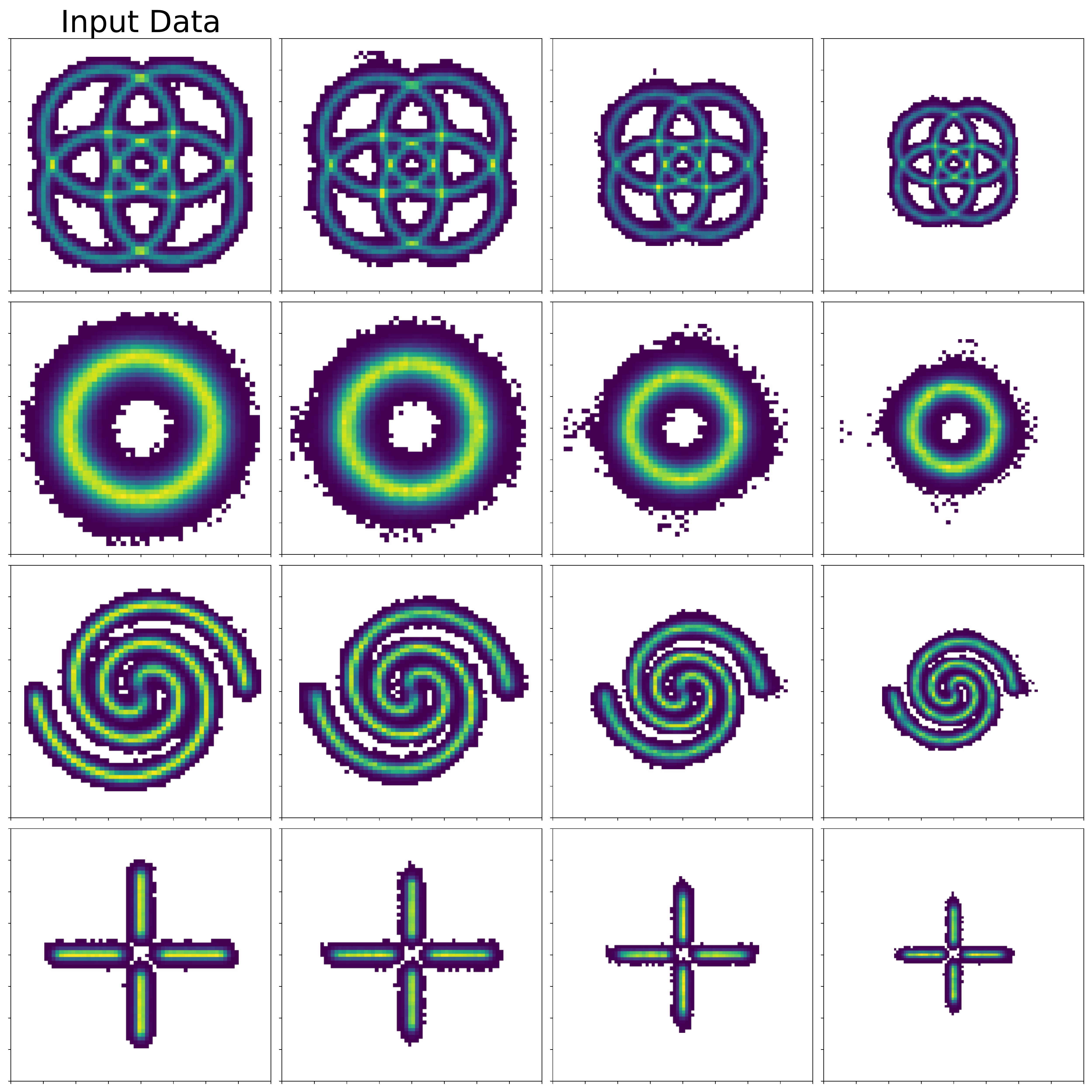}
    \caption{
        The conditional distributions produced by a flow for flow model.
    }
    \label{fig:radial_scale_grid}
\end{figure}

\begin{figure}[h]
    \centering 
    \includegraphics[width=\textwidth]{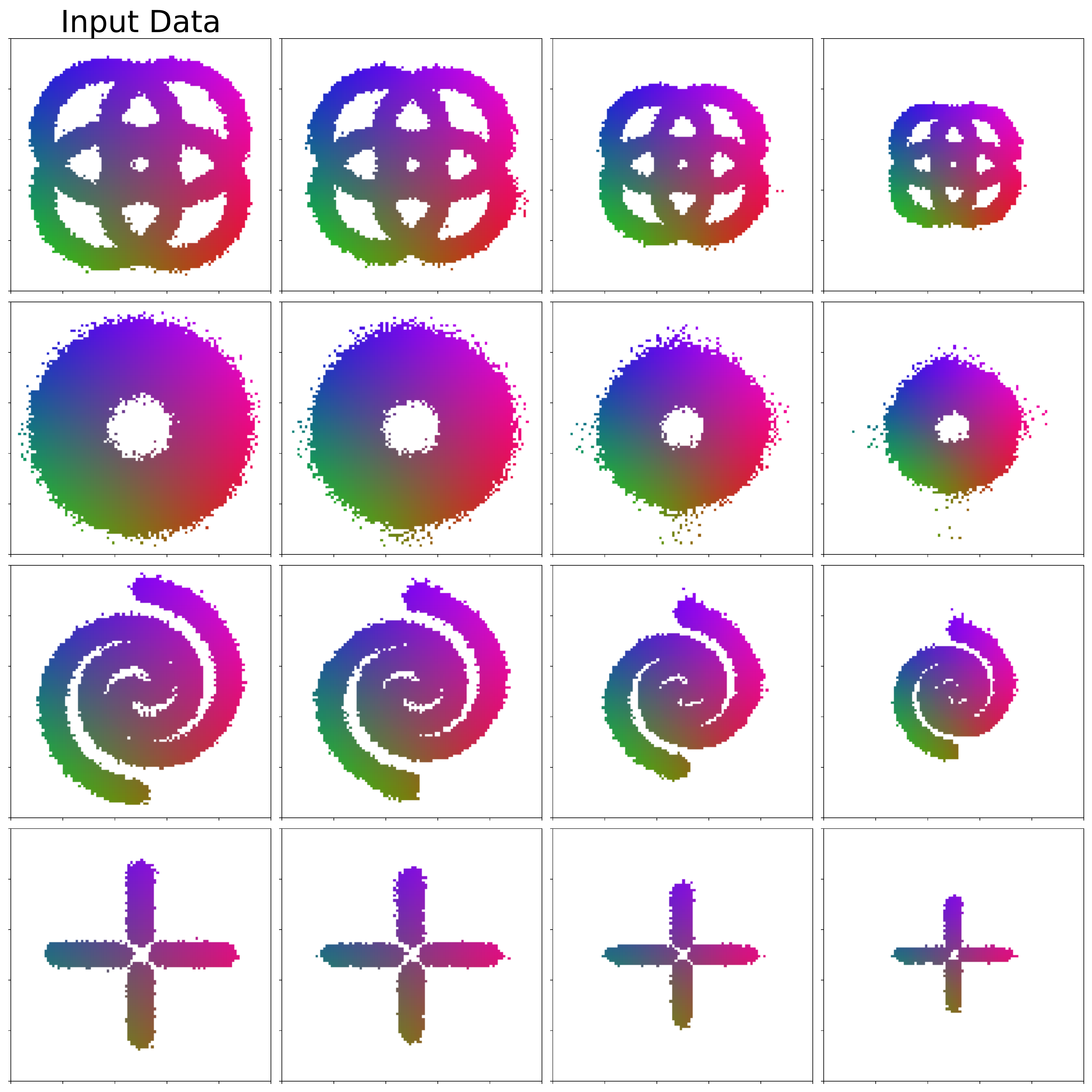}
    \caption{
        Color encodings of the maps produced by the flow for flow model without any regularisation. 
    }
    \label{fig:radial_scale_grid_colored}
\end{figure}

Similar behaviour as seen with rotating distributions can also be seen in conditional densities with radial scaling as shown in Fig.~\ref*{fig:radial_scale_grid}.
The minimality of the shifts learned by the flows for flows models is also consistent across these distributions as shown in Fig.~\ref*{fig:radial_scale_grid_colored}.

\end{document}